\documentclass[11pt,letterpaper]{article}
\usepackage{emnlp2016}
\usepackage{times}
\usepackage{latexsym}
\usepackage{multirow}
\usepackage{graphicx}
\usepackage{mathtools}
\usepackage{url}
\usepackage[font={normalsize}]{caption}

\usepackage{float}
\floatstyle{plaintop}
\restylefloat{table}
\emnlpfinalcopy

\def\emnlppaperid{***}

\title{Mining Compatible/Incompatible Entities from Question and Answering \\via Yes/No Answer Classification using Distant Label Expansion}

\author{Hu Xu$^1$, Lei Shu$^1$, Jingyuan Zhang$^1$ \and Philip S. Yu$^{1, 2}$\\
	$^1$Department of Computer Science, University of Illinois at Chicago, USA\\
	$^2$Tsinghua University, Beijing, China\\
	\{hxu48, lshu3, jzhan8, psyu\}@uic.edu\\
}

\date{}

\begin{document}

\maketitle

\begin{abstract}
Product Community Question Answering (PCQA) provides useful information about products and their features (aspects) that may not be well addressed by product descriptions and reviews. We observe that a product's \emph{compatibility} issues with other products are frequently discussed in PCQA and such issues are more frequently addressed in accessories, i.e., via a yes/no question ``Does this mouse work with windows 10?''. In this paper, we address the problem of extracting compatible and incompatible products from yes/no questions in PCQA. This problem can naturally have a two-stage framework: first, we  perform Complementary Entity (product) Recognition (CER) on yes/no questions; second, we identify the polarities of yes/no answers to assign the complementary entities a compatibility label (\textit{compatible}, \textit{incompatible} or \textit{unknown}). We leverage an existing unsupervised method for the first stage and a 3-class classifier by combining a distant PU-learning method (learning from positive and unlabeled examples) together with a binary classifier for the second stage. The benefit of using distant PU-learning is that it can help to expand more implicit yes/no answers without using any human annotated data. We conduct experiments on 4 products to show that the proposed method is effective.
\end{abstract}

\section{Introduction}
\label{intro}
\begin{figure}[ht] 
   \centering
   \includegraphics[width=3.2in]{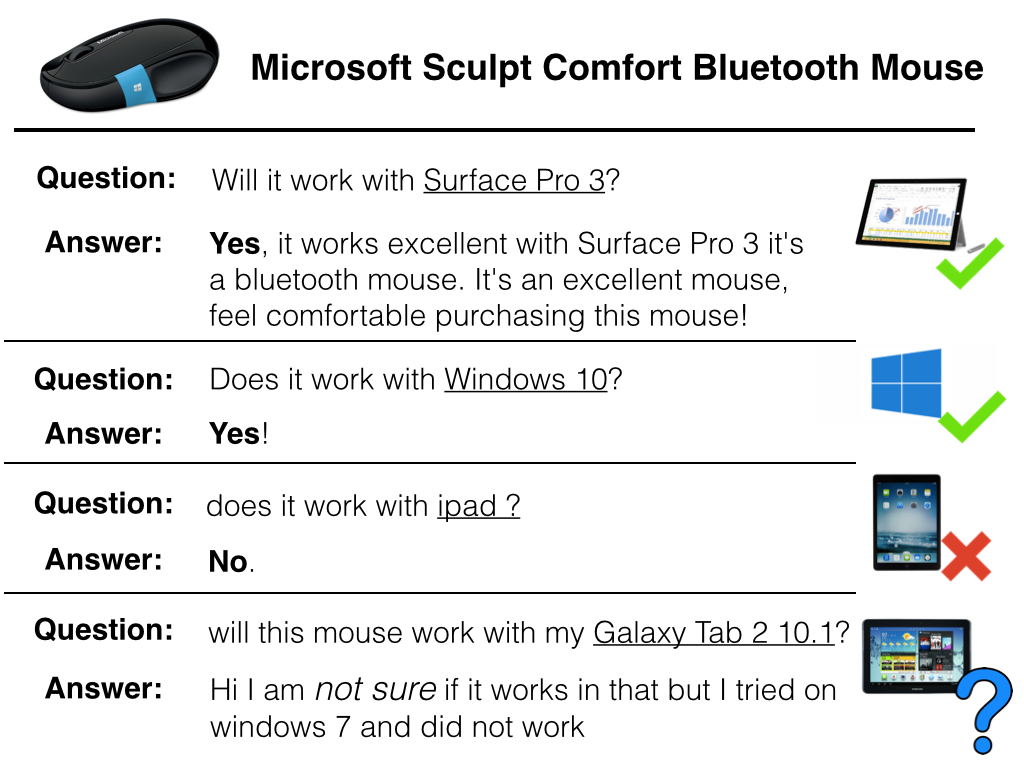}
   \caption{An example of QA pairs under a mouse product: complementary products are underlined in questions, ``Yes/No'' answers are bolded and ``Neutral'' answer is italicized; ``Microsoft Surface Pro 3'' and ``Windows 10'' are compatible products; ``iPad'' is an incompatible product; ``Samsung Galaxy Tab 2 10.0'' is unknown regarding compatibility issues.}
   \label{fig:sample}
\end{figure}

E-commerce websites like Amazon.com incorporate Product Community Question Answering (PCQA) into their websites to provide additional information about their products. Questions are usually posted by customers before their purchases and answers are provided by existing product owners or sellers. Compatibility issues are one popular topic in PCQA. As shown in Figure \ref{fig:sample}, one customer may write a question like ``Will it work with Surface Pro 3?''; existing customer may reply with ``Yes.''. From those 4 QA pairs discussing a Microsoft mouse, we know that the Microsoft mouse is compatible with  ``Microsoft Surface Pro 3'' and ``Windows 10'' but incompatible with ``iPad''. Furthermore, we have no idea whether ``Samsung Galaxy Tab 2 10.0'' is compatible or not with this mouse. Similar to our previous work in product reviews \cite{xu2016CER}, we call the mouse \emph{target entity} and those 4 products \emph{complementary entities} of the target entity. Each complementary entity forms a \emph{complementary relation} with the target entity. Each yes/no answer further assigns a \emph{compatibility label} to each complementary entity. 

Knowing which entity is compatible and which one is not is important because customers need to buy compatible ones and avoid incompatible ones. It is also important for manufacturers to realize the compatibility issues of their product. Further, recommender systems need to be aware of such issues and stay out of trouble of recommending incompatible products for their valued customers.

\textbf{Problem Statement}: We deal with the problem of identifying compatible and incompatible products from QA pairs in PCQA. More specifically, given a yes/no QA pair, we want to recognize complementary entities from questions and assign compatibility labels (\textit{compatible}, \textit{incompatible} or \textit{unknown}) to them according to the polarity (\textit{yes}, \textit{no} or \textit{neutral}) of the answers. 

We observe that compatibility issues are mostly discussed via yes/no questions rather than open questions. This is because customers tend to ask specific questions in PCQA. We leave the work of mining compatible/incompatible products on open questions to future work. Given the structure of a QA pair, our method naturally has a two-stage framework: Complementary Entity Recognition (CER) \cite{xu2016CER} and yes/no answer classification. For the first stage, we employ a similar approach as in \cite{xu2016CER}; for the second stage, it is reduced to a yes/no answer classification problem \cite{McAYan16}. We observe that the second stage provides further research opportunity since the polarities of many yes/no answers are implicit. For example, ``Will it work with Surface Pro 3? It works.'' has no explicit ``Yes'' but it is still a \textit{yes} answer. Therefore, exploiting implicit yes/no answers can further help to identify even more compatible/incompatible entities. 

To the best of our knowledge, there are no largely annotated implicit yes/no answers for PCQA. To save time-intensive annotation efforts, we leverage a distant PU-learning (learning from positive and unlabeled examples) method \cite{liu2003building,elkan2008learning} without using any human annotated answer. This is possible due to a simple observation: the beginning ``Yes'' or ``No'' word in explicit yes/no answers can serve as distant labels and can be used to expand implicit yes/no answers. For example, ``Yes, it works'' and ``It works'' have the same polarity. But the first answer is explicit and the second one is implicit. So the beginning word ``Yes'' can label ``Yes, it works'' as a \textit{yes} answer and further the implicit answer ``It works'' may also be labeled as a \textit{yes} answer due to its similarity with the former explicit \textit{yes} answer. 

Besides \textit{yes} and \textit{no} answers, we assume that there are also \textit{neutral} answers. For example, the last answer in Figure \ref{fig:sample} is a neutral answer and we have no obvious distant label for that type of answers. The framework of PU-learning (learning from positive and unlabeled examples) comes to rescue since it only requires positive examples and we already have many unlabeled answers. The idea of obtaining positive examples is simple: we leverage explicit answers (both \textit{yes} and \textit{no} answers) as positive examples and those explicit answers can expand to implicit answers via the PU-learning framework. Since all the explicit answers are distantly labeled, we have no human annotation effort at all. To further separate \textit{yes} and \textit{no} answers, we utilize a binary classifier trained from explicit \textit{yes/no} answers to classify all positive examples labeled by PU-learning.

The major contribution of this paper can be summarized as follows: we propose the problem of mining compatible/incompatible products from PCQA; we propose a two-stage framework to solve this problem without using any human annotated data. The rest of this paper is organized as follows: we describe related works in Section \ref{sec:rw}; In Section \ref{sec:r} and \ref{sec:yesno} we describe the proposed two-stage framework; we conduct experiments in Section \ref{sec:exp} and then draw our conclusion.

\section{Related Works}
\label{sec:rw}
The problem of Complementary Entity Recognition (CER) is first proposed by Xu et. al. \cite{xu2016CER}. However, our previous work focuses on product reviews and consider CER as a special kind of aspect extraction problem \cite{liu2015sentiment}. Determining the polarities of compatibility is reduced to a traditional sentiment classification problem. This paper focuses on yes/no QAs in PCQA and the polarities of compatibility is a yes/no answer classification problem. 

CER is closely related to entity recognition (e.g., Named Entity Recognition (NER) \cite{nadeau2007survey,zhou2002named} problem). The major differences are that many complementary entities are not named entities and CER heavily relies on the context of an entity (e.g., ``iPhone'' in ``I like my iPhone'' is not a complementary entity). Complementary entities are also studied as a social network problem in recommender systems \cite{zheng2009substitutes,McAPanLes15}. We discussed the benefit of CER over social network problem in \cite{xu2016CER} so we omit here but keep a performance comparison in Section \ref{sec:exp}.

Community Question and Answering (CQA) has been well studied in literature \cite{liu2008predicting,nam2009questions,li2010routing,anderson2012discovering}. More specifically, product Community Question and Answering (PCQA) is studied in \cite{McAYan16,liuretrieving}. They both try to find relevance between reviews and questions. \cite{McAYan16} takes questions from PCQA as queries and retrieve relevant reviews that can answer those queries. \cite{liuretrieving} considers questions in PCQA as summaries of reviews to help customers to identify relevant reviews. 

Extracting compatible/incompatible products from PCQA is very important. Based on our experience of annotating PCQA, we notice that PCQA usually addresses compatibility issues that are not well addressed by product description. This is because the number of complementary products for a target product can be unlimited so it is impractical to cover all of them. We also bring out the test dataset used in \cite{xu2016CER} for a comparison (Section \ref{sec:exp}). We notice that PCQA addresses compatibility issues in a different perspective compared to product reviews. PCQA tends to be specific on compatibility issues; reviews are free to talk about their experiences (e.g, opinions on features/aspects). For example, customers tend to ask more specific questions like ``Will it work with Surface Pro 3'' rather than ``Will it work with my tablet?'' since the latter question is pointless; reviews are typical datasets for opinion mining and aspects extraction \cite{liu2015sentiment}. Also, it is common to see general complementary products like ``It works with my tablet.'' in reviews since reviewers do not need to specify which tablet they have. 

Determining the polarity of a yes/no answer is closely related to answer summarization subtask B in SemEval-2015 Task 3 \cite{marquez2015semeval}. The proposed problem differs from this subtask B in that our problem indirectly utilizes the polarity of an answer to classify complementary entity rather than directly summarizes the usefulness of an answer to a question. McAuley et. al \cite{McAYan16} classifies the polarity of a PCQA answer by simply training an SVM on unigrams of labeled answers. From their predictions, we observe that they may only label explicit yes/no answers (e.g., answers begin with a ``Yes'' or ``No'') and put many implicit answers (e.g., ``I think it works.'' implies a \textit{yes} answer) as \textit{uncertain}. Identifying more implicit yes or no answer is crucial to the proposed problems since a complementary entity does not provide much information without its compatibility label (\textit{compatible}, \textit{incompatible} or \textit{uncertain}). 

The proposed method utilizes the PU-learning framework \cite{liu2003building,elkan2008learning}, which can be used to expand positive examples (both \textit{yes} and \textit{no} answers). We demonstrate that PU-learning can improve the recall by exploiting more implicit answers in Section \ref{sec:exp}.

\section{Two-stage Framework and CER}
In this section, we first introduce the two-stage framework of the proposed method. Then we briefly introduce the method for CER in \cite{xu2016CER}.

\subsection{Two-stage Framework}
Since complementary entities are mentioned in yes/no questions and their polarities of compatibility information are in answers, the proposed method naturally has a two-stage framework: \\
\textbf{Complementary Entity Recognition}: we extract complementary entities from questions using dependency paths almost the same as in \cite{xu2016CER}. It utilizes a large amount of unlabeled reviews under the same category as the target entity to expand knowledge about domain-specific verbs.\\
\textbf{Identifying Polarities of Yes/No Answers}: then we determine the polarity (\textit{yes}, \textit{no} or \textit{neutral}) of yes/no answers for each question with complementary entity and assign a compatibility label (\textit{compatible}, \textit{incompatible} or \textit{unknown}) to it. We form this 3-class classification via PU-learning and a binary SVM classifier in Section \ref{sec:yesno}.

\subsection{Complementary Entity Recognition}
\label{sec:r}
We briefly introduce the method used in \cite{xu2016CER} and how the dependency paths can be used in questions of PCQA (details of dependency paths can be found in the original paper). The basic idea is to use dependency paths to identify the context of complementary relations around complementary entities. Dependency paths can match dependency relations parsed through dependency parsing\footnote{We use Stanford CoreNLP (\url{http://stanfordnlp.github.io/CoreNLP/}) as our dependency parser}, which parses a sentence into a set of dependency relations. In our previous work, we notice that the verbs used to indicate a complementary relation can be unlimited and product specific. So we utilize another novel set of dependency paths that are in high precision but low recall to expand knowledge about complementary entities on a large amount of unlabeled review. We use similar ideas in this paper since verbs in questions of PCQA are also unlimited and product specific. But we do not incorporate candidate complementary entities into dependency paths when performing extractions because complementary entities are rather specific and diverse in PCQA and general entities are rarely mentioned. 

We still keep candidate complementary entities when expanding knowledge about domain-specific verbs. The knowledge expansion process is the same as our previous work. We start with seed verbs ``work'' and ``fit''. Then we first expand candidate complementary entities on the large unlabeled reviews. Then we use those candidate complementary entities to expand domain-specific verbs, e.g., ``insert'' for \textit{micro SD card} and ``hold'' for \textit{tablet stand}. The idea of using reviews rather than questions in PCQA to expand domain knowledge is that reviews contain a lot of the same general complementary entities (e.g. ``tablet'') that can easily appear in different reviews. However, ``Samsung Galaxy S6'' may be in low frequency in PCQA.

\section{Identifying the Polarities of Yes/No Answers}
\label{sec:yesno}
After CER, we need to identify whether a product is compatible or not with the target product. We assume a yes/no answer can clearly identify the polarities of the compatibility of a complementary entity for the target entity. We only classify the polarities of answers for successful extraction of complementary entities. 

\subsection{Motivations}
We assume that largely annotated \textit{yes} and \textit{no} answers are not available for training. We observe that the explicit mentions of ``Yes'' or ``No'' at the beginning of each answer are indicators of \textit{yes} or \textit{no} answers respectively. So they can be used for prediction directly. However, not every answer in PCQA begins with an explicit ``Yes'' or ``No'' word, but the polarity of the answer can still be implicitly expressed. For example, ``Yes, it works.'' and ``It works.'' have the same \textit{yes} polarity, but the latter answer does not have an explicit word ``Yes''. From the test data in Section \ref{sec:exp}, we observe that using explicit mentions of ``Yes'' or ``No'' contribute about 60\% of accuracy of \textit{yes} or \textit{no} answer classification. Without identifying those implicitly mentioned polarities, the polarities of compatibility for many complementary entities are uncertain. 

\subsection{Distant PU-Learning Classifier and Binary Classifier}
We distribute the classification task into 2 classifiers. First, we use PU learning to train a classifier that can separate \textit{yes} or \textit{no} answers from \textit{neutral} answers. Second, we train a \textit{yes} or \textit{no} binary classifier by using the explicit yes/no examples.

From the previous examples of ``Yes, it works.'' and ``It works.'', we observe that the beginning word ``Yes'' or ``No'' is optional for a \textit{yes} or \textit{no} answer respectively. So ``Yes'' can be served as a distant label for the training example ``It works''. We select all answers beginning with ``Yes'' or ``No'' as training examples and take the first words as distant labels and transform the remaining words of the answer to features. However, we notice that there is no obvious distant label for \textit{neutral} answers (e.g., ``I am not sure.''). Therefore, it is impossible to train a 3-class classifier directly.

Instead, we utilize PU-learning framework \cite{liu2003building,elkan2008learning} to first separate implicit yes/no answers from \textit{neutral} answers. PU-learning is a machine learning method using only positive and unlabeled examples (no negative examples are labeled). To get positive examples, we first combine all examples distantly labeled by ``Yes'' or ``No'' (the first word in an answer) together. Unlabeled examples can be easily collected from PCQA answers as long as the first word is not ``Yes'' or ``No''. Please note that unlabeled examples contain both implicit yes/no answers and \textit{neutral} answers. We utilize the implementation of PU learning described in \cite{liu2003building}. 

Lastly, we train a yes/no binary classifier using the same positive examples (explicit ``Yes'' or ``No'' answers) used in PU learning. But this time we separate the distant labels ``Yes'' and ``No''. By combining a PU-learning classifier and a binary classifier, we actually build a 3-class classifier for implicit yes/no answer classification.

During testing, we ensemble the results from the first ``Yes'' or ``No'' word prediction, the PU-learning classifier and the binary classifier together. We first detect whether the answer is an explicit yes/no answer by checking the first word. If the first word is a ``Yes'' (or ``No''), we output label \textit{yes} (or \textit{no}); otherwise we use PU-learning classifier to predict whether the answer is an implicit yes/no answer or \textit{neutral}; if it outputs negative, we consider the answer as \textit{neutral}; otherwise we consider it as an implicit yes/no answer and use the binary classifier to predict \textit{yes} or \textit{no}. We demonstrate this method using SVM as the base classifier for both the PU-learning classifier and the binary classifier in Section \ref{sec:exp}. In reality, other base classifiers can also be adopted.

\section{Experimental Results}
\label{sec:exp}
In this section, we first describe the dataset used for testing; then we introduce the evaluation methods and the baselines; lastly, we analyze the results.

\subsection{Dataset}
We crawl questions with at least one answers from product Community Question and Answering of Amazon.com and choose 4 products for test purpose. The 4 products are ``stylus'', ``micro SD card'', ``mouse'' and ``tablet stand''. We label complementary entities mentioned in each question and the answers as \textit{yes}, \textit{no} or \textit{neutral}. The whole test dataset is labeled by 3 annotators independently. The initial agreement is 93\%. Then disagreements are discussed and final agreements are reached among all annotators. To obtain knowledge about domain-specific verbs, we use 6000 reviews for each product similar as in \cite{xu2016CER}. We also select about 220 reviews for each product and label them in a similar way to show the difference between product QA community and reviews. The agreement for reviews is 82\%. The statistics of the datasets\footnote{The dataset will be available on the first author's website: \url{https://www.cs.uic.edu/~hxu/} } can be found in Table \ref{table:testingdata}. 

\begin{table*}
\centering
\scalebox{0.8}{
\begin{tabular}{ l | c | c | c | c | c | c | c | c | l }
\hline
Product & Q/R & Q/RSent. & CP & Density & Uniq. CP & Pos.& Neg. & Neu. & Examples\\
\hline
\multicolumn{9}{ l }{QA} \\
\hline
Stylus & 255 & 315 & 164 & \textbf{0.52} & \textbf{141} & 91 & 10 & 63  & 
\multirow{4}{*}{\begin{tabular}[t]{@{}l@{}} does it work with a nook? \\Will this work on a Samsung smart tv? \\Is this compatible with a HP Chromebook \\does it work with android \end{tabular} } \\
Micro SD Card & 277 & 352 & 223 & \textbf{0.63} & \textbf{200} & 162 & 16 & 45  \\
Mouse & 244 & 364 & 142 & \textbf{0.39} & \textbf{116} & 85 & 22 & 35  \\
Tablet Stand & 146 & 200 & 116 & \textbf{0.58} & \textbf{104} & 71 & 13 & 32  \\
\hline
\multicolumn{9}{ l }{Review} \\
\hline
Stylus & 216 & 892 & 165 & 0.18 & 83 & 130 & 32 & 3  & 
\multirow{4}{*}{\begin{tabular}[t]{@{}l@{}}It fits my phone well. \\It works with my tablet. \\Works with my phone. \end{tabular} } \\
Micro SD Card & 216 & 802 & 193 & 0.24 & 134 & 173 & 15 & 5  \\
Mouse & 216 & 1158 & 221 & 0.19 & 140 & 150 & 60 & 11  \\
Tablet Stand & 218 & 784 & 154 & 0.2 & 92 & 141 & 12 & 1  \\
\hline 
\end{tabular}
}
\caption{Statistics of test QA dataset and corresponding reviews: we list number of QA pairs and reviews (Q/R), number of sentences in questions and sentences in reviews (Q/RSent.), number of complementary product mentions (CP), number of complementary products per question sentence or review sentence (Density), number of unique complementary products (Uniq. CP), number of positive/negative/neutral complementary products mentions (Pos./Neg./Neu.) and a few example sentences.}
\label{table:testingdata}
\end{table*}

We observe that PCQA has higher densities (complementary products per sentence) of mentions of complementary entities. Further, PCQA has unique complementary entities since repeatedly asking the same question does not make sense. So identifying complementary entities from PCQA is much effective than that from customer reviews. Based on our experience of annotation, complementary entities mentioned in PCQA and in customer reviews are different. In PCQA, potential buyers frequently mention specific complementary entities as named entities (e.g., ``Microsoft Surface Pro 3'') to make their questions more accurate; in customer reviews, complementary entities can be general complementary products like ``tablet'', ``phone'', which is much less meaningful than specific products.

We also read the product descriptions of these 4 products and count the number of compatible products, including general products like ``Android tablets''. There are 13, 9, 5 and 55 compatible products for the stylus, micro SD card, mouse, tablet stand respectively. No incompatible products are mentioned in descriptions. So we can conclude that PCQA provides more information about compatibility issues.

\subsection{Compared Methods and Evaluation Methods}
We first perform separate evaluations on CER and yes/no answer classification. Then we combine those two stages together to evaluate the overall accuracy. For CER, we count true positive, false positive and false negative to compute precision $\mathcal{P}$, recall $\mathcal{R}$ and F1-score $\mathcal{F}_1$. We consider each question as an instance and the dependency paths are applied to each sentence in that question and then the extractions combined to form one prediction. A prediction contributes to one count of true positive when the extracted complementary products match the labeled complementary entities in one question; more or less predicted complementary entities in one question are treated as one count of false positive; failed extraction from one question is treated as one count of false negative.\\
\textbf{Noun Phrase Chunker}: Most of the product names mentioned in questions of PCQA are noun phrases, we use the same noun phrase chunking pattern as the proposed method to extract noun phrases directly from questions and take them as complementary products.\\
\textbf{UIUC NER}: We use UIUC Named Entity Tagger \cite{ratinov2009design} to perform Named Entity Recognition (NER) on questions in PCQA. UIUC NER has 18 labels in total and we consider words or phrases labeled as ``PRODUCT'' or ``ORG'' as predictions of complementary products.\\
\textbf{Sceptre}: We also retrieve the top 25 complements for the same 4 products from \cite{McAPanLes15}'s Sceptre and adapt their results for a comparison. Direct comparison is impossible since they deal with a link prediction problem and consider ``Items also bought'' as complementary products for training/testing. We label and compute the precision for the top 25 predictions and assume annotators have the same background knowledge for both their datasets and ours. We observe that the predicted complementary products are irrelevant products like ``network cables'', ``mother board'', etc. and all 4 products have similar complementary products. We mostly consider ``Windows'' as complementary products for ``Mouse''.\\
\textbf{CER6K}: This method is the method proposed in \cite{xu2016CER}. Specifically, it uses 6000 reviews to expand domain-specific verbs.

\begin{table*}[htp]
\centering
\scalebox{0.95}{
\begin{tabular}{ l | c c c | c c c | c | c c c }
\hline
\multirow{2}{*}{Product} & 
\multicolumn{3}{ |c| }{NP Chunker} &
\multicolumn{3}{ |c| }{UIUC NER} &
\multicolumn{1}{ |c| }{Sceptre} &
\multicolumn{3}{ |c }{CER6K} \\
\cline{2-11} & 
$\mathcal{P}$ & $\mathcal{R}$ & $\mathcal{F}_1$ &
$\mathcal{P}$ & $\mathcal{R}$ & $\mathcal{F}_1$ & 
$\mathcal{P}$@25 & 
$\mathcal{P}$ & $\mathcal{R}$ & $\mathcal{F}_1$ \\
\hline
Stylus  &  0.599 & 0.774 & 0.676 & 0.931 & 0.329 & 0.486 & 0.04 & 0.917 & 0.805 & \textbf{0.857} \\
Micro SD Card  &  0.734 & 0.632 & 0.68 & 0.843 & 0.336 & 0.481 & 0.16 & 0.973 & 0.798 & \textbf{0.877} \\
Mouse  &  0.498 & 0.704 & 0.583 & 0.842 & 0.225 & 0.356 & 0.16 & 0.92 & 0.725 & \textbf{0.811} \\
Tablet Stand  &  0.723 & 0.629 & 0.673 & 0.968 & 0.259 & 0.408 & 0.04 & 0.949 & 0.647 & \textbf{0.769} \\
\hline 
\end{tabular}
}

\caption{Different methods for CER in precision, recall and F1-score}
\label{table:comparison}
\end{table*}

\begin{table*}[htp]
\centering
\scalebox{0.9}{
\begin{tabular}{ l | c | c | c | c | c ||| c }
\hline
Product & 
Yes/No &
Sentiment Parser &
One-Class SVM &
3-Class SVM &
PU SVM & 
Overall Results \\
\hline
Stylus & 0.646 & 0.524 & 0.567 & 0.652 & \textbf{0.72} & \textbf{0.749} \\
Micro SD Card  &  0.673 & 0.646 & 0.7 & 0.682 & \textbf{0.776} & \textbf{0.755} \\
Mouse  &  0.606 & 0.535 & 0.641 & 0.634 & \textbf{0.69} & \textbf{0.738} \\
Tablet Stand  &  0.569 & 0.517 & 0.595 & 0.586 & \textbf{0.802} & \textbf{0.692} \\
\hline 
\end{tabular}
}
\caption{Comparison of accuracy of answer classification: the last column is the combined results of CER and answer classification.}
\label{table:opcomparison}
\end{table*}

Next, we perform a separate evaluation on yes/no answer classification. We assume the accuracies of complementary entities extraction are 100\% and errors do not affect answer classification. We only classify answers to questions that have labeled complementary entities.\\
\textbf{Yes/No}: This simple baseline predicts the polarities of yes/no answers based on the first ``Yes'' or ``No'' word in an answer; if the first word is not ``Yes'' or ``No'', it predicts the answer as \textit{neutral}.\\
\textbf{Sentiment Parser}: We utilize the RNN-based sentiment parser \cite{socher2013recursive} to get the sentiment polarities of the first sentences in answers. We observe that opinions expressed in answers can indicate the polarities of answers. For example, ``It works well.'' indicates a positive answer. We use the results of sentiment parsing to get more implicit yes/no answers and combine the explicit answers outputted by Yes/No baseline.\\
\textbf{One-Class SVM(Bigram)}: Similar to PU learning, one-class SVM is also a classifier without using negative training examples. But one-class SVM does not use unlabeled data during the training process. This means the \textit{neutral} answers are only available in testing. We feed one-class SVM with 20000 explicit yes/no answers as training examples. We utilize Scikit Learn\footnote{http://scikit-learn.org/} as the implementation of one-class SVM. Then similar to the proposed method, we train a yes/no binary SVM classifier and pipeline Yes/No, One-Class SVM and binary yes/no classifier together.\\
\textbf{3-Class SVM(Bigram)}: We train a 3-class SVM classifier using the answer predictions from \cite{McAYan16}. Their method originally uses 1000 labeled data as the training data for answer prediction. Since the labeled training data is not available, we use their predictions as the training data. We select 4000 examples for each class as training examples and ensemble the results with Yes/No baseline.\\
\textbf{PU SVM(Bigram)}: This is described in Section \ref{sec:yesno}. We use 3000 \textit{yes} answers and 3000 \textit{no} answers as positive examples and 6000 unlabeled answers. We use bigrams as features for prediction and PU learning method in \cite{liu2003building} as the implementation of PU learner.\\
Finally, we combine the results of CER6K and PU SVM to get the Overall Results.

\subsection{Result Analysis}
\textbf{CER}: From Table \ref{table:comparison}, we can see that CER6K performs the best. NP chunker performs better than UIUC NER because NER heavily relies on capital letters as features but PCQA tends to have typos in lower case (e.g., ``samsung'' instead of ``Samsung''). The precision of Sceptre is low because the ``Items also bought'' training data tend to be noisy for accessories. We further observe that the recall of ``tablet stand'' is relatively low. We examine the data and find that many errors are due to parsing errors (e.g., the POS tagger treats ``stand'' as a verb, which makes dependency parsing incorrect.).\\
\textbf{Yes/No Answer Classification}: In Table \ref{table:opcomparison}, we compare the results of yes/no answer classification. The first 5 methods are performed only on the answers with questions that have human-annotated complementary entities. The last column is the combined results of CER6K and PU SVM(Bigram). All numbers are accuracies of classification for \textit{yes}, \textit{no} and \textit{neutral}. Given many explicit yes/no answers, the Yes/No baseline performs relatively good. Sentiment parser performs worse than the Yes/No baseline. We examine the results and find that sentiment parser tends to produce more errors on negative opinions. 3-class SVM(Bigram) does not have much improvement over Yes/No baseline. This is because \cite{McAYan16}'s predictions are mostly explicit \textit{yes} or \textit{no} answers. We guess they mostly label implicit \textit{yes} or \textit{no} answers as \textit{neutral}. PU learning performs better than One-class SVM because PU learning also leverages unlabeled data, even though the size of training data is smaller. In the overall results, we achieve accuracy around 70\%.

\section{Conclusions}
In this paper, we propose the problem of mining compatible and incompatible products from product Community Question and Answering (PCQA). We propose a two-stage framework to solve this problem. We first extract complementary entities from each question using a dependency rule-based method; then we determine the labels of compatibility for complementary entities from the polarities of yes/no answers. We leverage a distant PU learning method to identify extra implicit polarities of yes/no answers without using any human-labeled training data. Experiments show that the proposed method can exploit more implicit answers.

\section*{Acknowledgment}
This work is supported in part by NSF through grants IIS-1526499 and CNS-1626432. We gratefully acknowledge the support of NVIDIA Corporation with the donation of the Titan X GPU used for this research.

\bibliography{HU_XU}

\begin{thebibliography}{}

\bibitem[\protect\citename{Anderson \bgroup et al.\egroup
  }2012]{anderson2012discovering}
Ashton Anderson, Daniel Huttenlocher, Jon Kleinberg, and Jure Leskovec.
\newblock 2012.
\newblock Discovering value from community activity on focused question
  answering sites: a case study of stack overflow.
\newblock In {\em Proceedings of the 18th ACM SIGKDD international conference
  on Knowledge discovery and data mining}, pages 850--858. ACM.

\bibitem[\protect\citename{Elkan and Noto}2008]{elkan2008learning}
Charles Elkan and Keith Noto.
\newblock 2008.
\newblock Learning classifiers from only positive and unlabeled data.
\newblock In {\em Proceedings of the 14th ACM SIGKDD international conference
  on Knowledge discovery and data mining}, pages 213--220. ACM.

\bibitem[\protect\citename{Li and King}2010]{li2010routing}
Baichuan Li and Irwin King.
\newblock 2010.
\newblock Routing questions to appropriate answerers in community question
  answering services.
\newblock In {\em Proceedings of the 19th ACM international conference on
  Information and knowledge management}, pages 1585--1588. ACM.

\bibitem[\protect\citename{Liu \bgroup et al.\egroup }2003]{liu2003building}
Bing Liu, Yang Dai, Xiaoli Li, Wee~Sun Lee, and Philip~S Yu.
\newblock 2003.
\newblock Building text classifiers using positive and unlabeled examples.
\newblock In {\em Data Mining, 2003. ICDM 2003. Third IEEE International
  Conference on}, pages 179--186. IEEE.

\bibitem[\protect\citename{Liu \bgroup et al.\egroup }2008]{liu2008predicting}
Yandong Liu, Jiang Bian, and Eugene Agichtein.
\newblock 2008.
\newblock Predicting information seeker satisfaction in community question
  answering.
\newblock In {\em Proceedings of the 31st annual international ACM SIGIR
  conference on Research and development in information retrieval}, pages
  483--490. ACM.

\bibitem[\protect\citename{Liu \bgroup et al.\egroup }2016]{liuretrieving}
Mengwen Liu, Yi~Fang, Dae~Hoon Park, Xiaohua Hu, and Zhengtao Yu.
\newblock 2016.
\newblock Retrieving non-redundant questions to summarize a product review.
\newblock pages 385--394.

\bibitem[\protect\citename{Liu}2015]{liu2015sentiment}
Bing Liu.
\newblock 2015.
\newblock {\em Sentiment Analysis: Mining Opinions, Sentiments, and Emotions}.
\newblock Cambridge University Press.

\bibitem[\protect\citename{M{\`a}rquez \bgroup et al.\egroup
  }2015]{marquez2015semeval}
Llu{\'\i}s M{\`a}rquez, James Glass, Walid Magdy, Alessandro Moschitti, Preslav
  Nakov, and Bilal Randeree.
\newblock 2015.
\newblock Semeval-2015 task 3: Answer selection in community question
  answering.
\newblock In {\em Proceedings of the 9th International Workshop on Semantic
  Evaluation (SemEval 2015)}.

\bibitem[\protect\citename{McAuley and Yang}2016]{McAYan16}
J.~McAuley and A.~Yang.
\newblock 2016.
\newblock Addressing complex and subjective product-related queries with
  customer reviews.
\newblock In {\em World Wide Web}.

\bibitem[\protect\citename{McAuley \bgroup et al.\egroup }2015]{McAPanLes15}
J.~J. McAuley, R.~Pandey, and J.~Leskovec.
\newblock 2015.
\newblock Inferring networks of substitutable and complementary products.
\newblock In {\em KDD}.

\bibitem[\protect\citename{Nadeau and Sekine}2007]{nadeau2007survey}
David Nadeau and Satoshi Sekine.
\newblock 2007.
\newblock A survey of named entity recognition and classification.
\newblock {\em Lingvisticae Investigationes}, 30(1):3--26.

\bibitem[\protect\citename{Nam \bgroup et al.\egroup }2009]{nam2009questions}
Kevin~Kyung Nam, Mark~S Ackerman, and Lada~A Adamic.
\newblock 2009.
\newblock Questions in, knowledge in?: a study of naver's question answering
  community.
\newblock In {\em Proceedings of the SIGCHI conference on human factors in
  computing systems}, pages 779--788. ACM.

\bibitem[\protect\citename{Ratinov and Roth}2009]{ratinov2009design}
Lev Ratinov and Dan Roth.
\newblock 2009.
\newblock Design challenges and misconceptions in named entity recognition.
\newblock In {\em Proceedings of the Thirteenth Conference on Computational
  Natural Language Learning}, pages 147--155. Association for Computational
  Linguistics.

\bibitem[\protect\citename{Socher \bgroup et al.\egroup
  }2013]{socher2013recursive}
Richard Socher, Alex Perelygin, Jean~Y Wu, Jason Chuang, Christopher~D Manning,
  Andrew~Y Ng, and Christopher Potts.
\newblock 2013.
\newblock Recursive deep models for semantic compositionality over a sentiment
  treebank.
\newblock In {\em Proceedings of the conference on empirical methods in natural
  language processing (EMNLP)}, volume 1631, page 1642. Citeseer.

\bibitem[\protect\citename{Xu \bgroup et al.\egroup }2016]{xu2016CER}
Hu~Xu, Sihong Xie, Lei Shu, and Philip~S. Yu.
\newblock 2016.
\newblock Cer: Complementary entity recognition via knowledge expansion on
  large unlabeled product reviews.
\newblock In {\em Proceedings of IEEE International Conference on Big Data}.

\bibitem[\protect\citename{Zheng \bgroup et al.\egroup
  }2009]{zheng2009substitutes}
Jiaqian Zheng, Xiaoyuan Wu, Junyu Niu, and Alvaro Bolivar.
\newblock 2009.
\newblock Substitutes or complements: another step forward in recommendations.
\newblock In {\em Proceedings of the 10th ACM conference on Electronic
  commerce}, pages 139--146. ACM.

\bibitem[\protect\citename{Zhou and Su}2002]{zhou2002named}
GuoDong Zhou and Jian Su.
\newblock 2002.
\newblock Named entity recognition using an hmm-based chunk tagger.
\newblock In {\em proceedings of the 40th Annual Meeting on Association for
  Computational Linguistics}, pages 473--480. Association for Computational
  Linguistics.

\end{thebibliography}
\bibliographystyle{emnlp2016}

\end{document}